\documentclass[runningheads]{llncs}
\usepackage{graphicx}
\usepackage[utf8]{inputenc} 
\usepackage[T1]{fontenc}    
\usepackage{hyperref}       
\usepackage{url}            
\usepackage{booktabs}       
\usepackage{amsfonts}       
\usepackage{nicefrac}       
\usepackage{microtype}      
\usepackage{lipsum}
\usepackage{amsmath}
\usepackage[ruled,vlined]{algorithm2e}
\usepackage{rotating}

\begin{document}
\title{ALIME: Autoencoder Based Approach for Local Interpretability}
%
%
\author{
    Sharath M. Shankaranarayana \and
    Davor Runje
}
\institute{
    ZASTI.AI\\
}
\maketitle              
\begin{abstract}
Machine learning and especially deep learning have garnered tremendous popularity in recent years due to their increased performance over other methods. The availability of large amount of data has aided in the progress of deep learning. Nevertheless, deep learning models are opaque and often seen as black boxes. Thus, there is an inherent need to make the models interpretable, especially so in the medical domain. In this work, we propose a locally interpretable method, which is inspired by one of the recent tools that has gained a lot of interest, called local interpretable model-agnostic explanations (LIME). LIME generates single instance level explanation by artificially generating a dataset around the instance (by randomly sampling and using perturbations) and then training a local linear interpretable model. One of the major issues in LIME is the instability in the generated explanation, which is caused due to the randomly generated dataset. Another issue in these kind of local interpretable models is the local fidelity. We propose novel modifications to LIME by employing an autoencoder, which serves as a better weighting function for the local model. We perform extensive comparisons with different datasets and show that our proposed method results in both improved stability, as well as local fidelity.

\end{abstract}
\section{Introduction and related work}

Machine learning models with good predictability, such as deep neural networks, are difficult to interpret, opaque, and hence considered to be black box models. Deep neural networks have gained significant prominence in healthcare domain and are increasingly being used in critical tasks such as disease diagnosis, survival analysis, etc. As a result, there is a pressing need to understand these models to ensure that they are correct, fair, unbiased, and/or ethical.

Currently, there is no precise definition of interpretability, and the definitions tend to depend upon the application. Some of the most common types of model explanations are \cite{int_book}: 
\begin{enumerate}
    \item \textbf{Example-based}. In this method, one is interested in knowing the point in the training set that has close resemblance with the test point one is interested in explaining \cite{example1,example2}.
    \item \textbf{Local}. In these methods, one is interested in deriving explanations by fitting an interpretable model locally around the test instance of interest \cite{lime}.
    \item \textbf{Global}. In these methods, one is interested in knowing the overall behaviour of the model. Global explanations attempt to understand underlying patterns in the model behaviour, usually by employing a series of rules for explanations as  \cite{global1,glaobal2}. 
\end{enumerate}
The interpretable and explainable methods can also be grouped based on other criteria, such as \emph{i)} Model agnostic or model specific \emph{ii)} Intrinsic or post-hoc \emph{iii)} Perturbation or saliency based, etc. 
Recently, post-hoc explainable methods, such as LIME \cite{lime}, have gained a lot of interest since $a$ $posteriori$ explanations may currently be the only option for explaining already trained models. These methods are model-agnostic and hence do not require understanding of the inner workings of trained model.  

Some of the desired characteristics of interpretable models are consistency and stability of explanations, and local fidelity or faithfulness of the model locally. Local post-hoc methods, such as LIME, lack in this regard. In this paper, we propose a modification to the popular framework for generating local explanations LIME\cite{lime} that improves both \emph{stability} and \emph{local fidelity} of explanations. 


Since our focus is on post-hoc and local interpretable methods, we restrict our literature review to include only those methods. As mentioned above, LIME\cite{lime} is one the most popular methods in local interpretability. In LIME, artificial data points are first generated using random perturbation around the instance of interest to be explained, and then a linear model is fit around the instance. The same authors extended LIME to include global explanations in \cite{glaobal2}. Most of the modifications to LIME have been in the line of selecting appropriate kind of data for training the local interpretable model. In \cite{related_h2o}, the authors use K-means clustering to partition the training dataset and use them for local models instead of perturbation-based data generation. In another work called DLIME \cite{related_dlime}, the authors employ Hierarchical Clustering (HC) for grouping the training data and later use one of the clusters nearest to the instance for training the local interpretable model. These modifications are aimed at addressing the lack of "stability", which is one of the serious issues in the local interpretable methods. The issue of stability occurs in LIME because of the data generation using random perturbation. Suppose we select an instance to be explained, the LIME generates different explanations (local interpretable models with different feature weights) at every run for the same instance. The works such as \cite{related_dlime} use clusters from the training data itself to address this problem. Since the samples are always taken from the same cluster of the training set, there is no variability in the feature weights at different runs for a particular instance.

Using the training set would make the black box model (the model for which we seek explanations) overfit by creating a table where each individual example in the training set is assigned to its class (i.e. exact matching of training instances), Therefore, using these results in training the local interpretable model would produce incorrect results, since the explanation instance of interest is from the test set and we do not have information as to how the black box model behaves upon encountering new data points as present in the test set. Thus, we ask the following question:

\begin{quote}
    \emph{Can we improve the stability of local interpretable model while sampling randomly generated points?}
\end{quote}
In other words, we wish to see if we can still improve the stability by following the same sampling paradigm of LIME and therefore not using the training set for local interpretable model as done in \cite{related_h2o,related_dlime}.   

Another important issue in local intepretable methods is \emph{locality}, which refers to the the neighbourhood around which the local surrogate is trained and in \cite{related_locality}, the authors show that it is non-trivial to define the right neighbourhood and how it could impact the \emph{local fidelity} of the surrogate. A straightforward way to improve the stability is to simply generate a large set of points and use it to train the local surrogate. Although doing so would improve stability, it also decreases local fidelity or local accuracy. Thus, we ask another question:
\begin{quote}
\emph{Can we improve the stability while simultaneously maintaining the local fidelity?} 
\end{quote}

In this paper, we mainly focus on answering the questions above by introducing an autoencoder-based local interpretability model ALIME. Our contributions can be summarized as follows: 
\begin{itemize}
    \item we propose a novel weighting function as a modification to LIME to address the issues of stability and local fidelity, and
    \item we perform extensive experiments on three different healthcare datasets to study the effects and compare with LIME.
\end{itemize}

\section{Methods}

Since our model builds upon LIME, we begin by a short introduction to LIME, and then describe our proposed modifications.

\subsection{LIME}

Local surrogate models use interpretable models (such as ridge regression) in order to explain individual instance predictions of an already trained machine learning model, which could be a black box model. Local interpretable model-agnostic explanations (LIME) is a very popular recent work where, instead of training a global surrogate model, LIME trains a local surrogate model for individual predictions. LIME method generates a new dataset by first sampling from a distribution and later perturbing the samples. The corresponding predictions on this generated dataset given by the black box model are used as ground truth. On these pairs of generated samples and the corresponding black box predictions, an interpretable model is trained around the point of interest by weighting the proximity of the sampled instances to it. This new learned model has the constraint that it should be a good approximation of the black box model predictions locally, but it does not have to be a good global approximation.
Formally, a local surrogate model with interpretability constraint is written as follows:
\begin{equation}
    \textrm{explanation}(x) = \arg \min_{g \in G} L\left(f, g, \pi_x\right) + \Omega(g)
\end{equation}

The explanation model for instance $x$ is the model $g$ (e.g. linear regression model) minimizing loss $L$ (e.g. mean squared error), a measure of how close the explanation is to the prediction of the original model $f$ (e.g. a deep neural network model). The model complexity is denoted by $\Omega(g)$. $G$ is the family of possible explanations which, in our case, is a linear ridge regression model. The proximity measure $\pi_x$ defines how large the neighborhood around instance x is that we consider for the explanation. In practice, LIME only optimizes the loss part. 
The algorithm for training local surrogate models is as follows:
\begin{itemize}
    \item Select the instance of interest for which an explanation is desired for a black box machine learning model.
    \item Perturb the dataset and use the black box to make predictions for these new points.
    \item Weight the new samples according to their proximity to the instance of interest by employing some proximity metric, such as euclidean distance.
    \item Train a weighted, interpretable linear model, such as ridge regression, on the dataset.
    \item Explain the prediction by interpreting the local linear model by analyzing the coefficients of the local linear model.
\end{itemize}

\begin{algorithm}[H]
\SetAlgoLined
\DontPrintSemicolon
\SetKwInOut{Input}{Input}
\SetKwInOut{Output}{Output}
\SetKwComment{Comment}{\# }{}

\Input{Dataset $D_{train}$ with $K$ features, model $M$, instance $x$, number of points sampled $m$
}

\Output{feature importance at $x$}

\Begin{
\Comment{Sample new dataset from Gaussian distribution}
$D_{sample} \stackrel{\mathrm{m}}{\longleftarrow} \mathbf{Gaussian}(K)$  \;
\Comment{For all sampled points calculate:}
\ForEach{$y \in D_{sample}$}{
    \Comment{euclidean distance from $x$}
    $d(y)  \longleftarrow |y - x |$ \;
    \Comment{weight}
    $W(y) \longleftarrow e^{-d(y)}$ \;
}
\Comment{Fit linear model}
$L \longleftarrow \mathbf{LinearModel.fit}(D, W)$  \;
\Comment{Return weights of the linear model}
\Return{$L_w$}\;
}

\label{alg:lime}
\caption{LIME}
\end{algorithm}

\begin{figure}
\includegraphics[width=\textwidth]{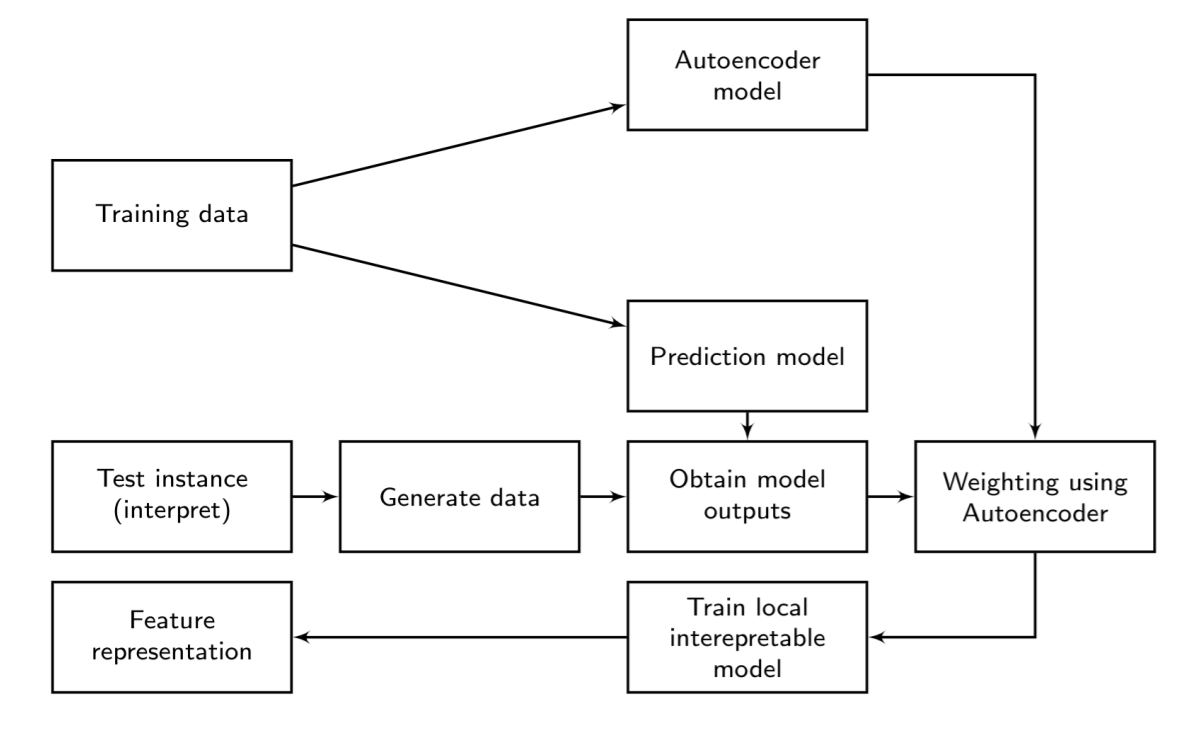}

\caption{Block diagram depicting our overall approach in ALIME} \label{fig:flowchart}
\end{figure}

\subsection{ALIME}
The high-level block diagram of our approach is shown in Figure \ref{fig:flowchart} and also described in Algorithm \ref{alg:alime}. Once the black box model is trained, we need to train a local interpretable model. Our first focus is on improving stability of the local interpretable model. For this, instead of generating data by perturbation every time for explaining an instance (as done in LIME), we generate a large number of data points beforehand by sampling from a Gaussian distribution. This has an added advantage, as we reduce the time-complexity by reducing the sampling operations. However, since we need to train a local model, we must ensure that for a particular instance, only the generated data around the instance is used for training the interpretable model.  
For this we use an autoencoder \cite{ae,denoising_ae} and thus, the most important change comes from the introduction of autoencoder. 

Autoencoder \cite{ae} is a neural network used to compress high dimensional data into latent representations. It consists of two parts: an encoder and a decoder. The encoder learns to map the high dimensional input space to a latent vector space, and the decoder maps the latent vector space to the original uncompressed input space. We use a variant of autoencoder called denoising autoencoder \cite{denoising_ae}, where the input is corrupted by adding a small amount of noise and then trained to reconstruct the uncorrupted input. Looked in another way, denoising is used as a proxy task to learn latent representations. We train an autoencoder with the help of the training data to be used for building the black box model. We first standardize the training data and then corrupt the training data by adding a small amount of additive white Gaussian noise. Then, the autoencoder is made to reconstruct the uncorrupted version of the input using the standard $L_2$ loss. Once trained, we employ the autoencoder as a weighting function, i.e., instead of computing the euclidean distances for the generated data and the instance to be explained on the original input space, we compute the distance on the latent vector space. For this, we compute the latent embeddings for all the generated points and the instance to be explained, and compute the distance on the embedded space. We discard the points with a distance larger than a predefined threshold and then, for the selected data points, we weight the points by using an exponential kernel as a function of distance. This way, we ensure locality and, since the autoencoders have been shown to better learn the manifold of data, it also improves local fidelity.    

\begin{algorithm}[H]
\SetAlgoLined
\DontPrintSemicolon
\SetKwInOut{Input}{Input}
\SetKwInOut{Output}{Output}
\SetKwComment{Comment}{\# }{}

\Input{Dataset $D_{train}$ with $K$ features, model $M$, instance $x$, number of points used $n$, number of points sampled $m$
}

\Output{feature importance at $x$}

\Comment{Precompute embeddings using autoencoder}
\Begin{
\Comment{Train autoencoder model}
$AE \longleftarrow \mathbf{AutoEncoder.fit}(D_{train})$  \;
\Comment{Sample new dataset from Gaussian distribution}
$D_{sample} \stackrel{\mathrm{m}}{\longleftarrow} \mathbf{Gaussian}(K)$  \;
\Comment{Calculate embeddings for $D_{sample}$}
$E \longleftarrow [AE(y)\ |\ y \in D_{sample} ]$ \;
}

\Comment{Given $x$, calculate feature importance}
\Begin{
\Comment{Calculate distances from $x$ in embedded space}
$d  \longleftarrow |E - E(x)|$ \;
\Comment{Find $n$-th minimum distance in $d$}
$d_{min} \longleftarrow \mathbf{min}(n, d)$ \;
\Comment{Collect $n$ closest points into a local dataset}
$D \longleftarrow [y \in D_{train} \textrm{ such that } |d_x - d_y| < d_{min} ]$\;
\Comment{Calculate weights}
$W \longleftarrow [ e^{-y}\ |\ y \in D]$ \;
\Comment{Fit linear model}
$L \longleftarrow \mathbf{LinearModel.fit}(D, W)$  \;
\Comment{Return weights of the linear model}
\Return{$L_w$}\;
}

\label{alg:alime}
\caption{ALIME}
\end{algorithm}

\section{Experiments and Results}

For the sake of experiments, we use three datasets belonging to healthcare domain from the UCI repository\cite{uci}:
\begin{enumerate}
    \item Breast Cancer dataset \cite{breast_cancer}. A widely used dataset that consists of 699 patient observations and 11 features used to study breast cancer.
    \item Hepatitis Patients dataset \cite{hep}. Dataset consisting of 20 features and 155 patient observations.
    \item Liver Patients dataset \cite{liverdisease}. Indian liver patient dataset consisting of 583 patient observations and 11 features used to study liver disease.
\end{enumerate}
\begin{figure}[tp!]
\includegraphics[width=\textwidth]{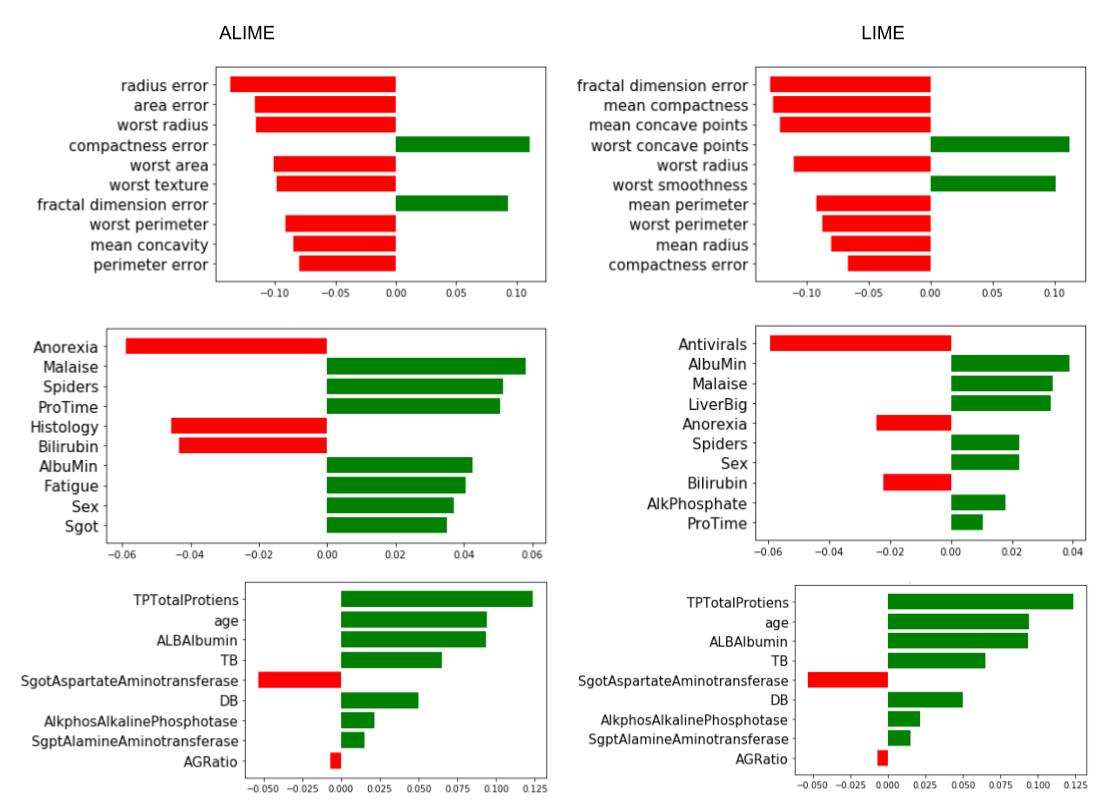}
%
\caption{Comparisons between our method and LIME \cite{lime}. The first row corresponds to the Breast Cancer dataset, the second row corresponds to the Hepatitis Patients dataset and the third row corresponds to the Liver Patients dataset} \label{fig:results}
\end{figure}

As a black box model, we train a simple feed forward neural network with a single hidden layer having $30$ neurons and $2$ neurons in the output layer for the two classes, and train the network using binary cross entropy loss. For all the three datasets, we use $70-30$ split for training and testing. We obtained accuracies of $0.95$, $0.87$ and $0.83$ on the above mentioned three datasets respectively.   
The sample results for instances from the three datasets are shown in Figure \ref{fig:results}. The red bars in the figure show the negative coefficients, and green bars show the positive coefficients of the linear regression model. The positive coefficients show the positive correlation among the dependent and independent attributes. On the other hand, negative coefficients show the negative correlation among the dependent and independent attributes. 

\begin{figure}[tp!]
\includegraphics[width=\textwidth]{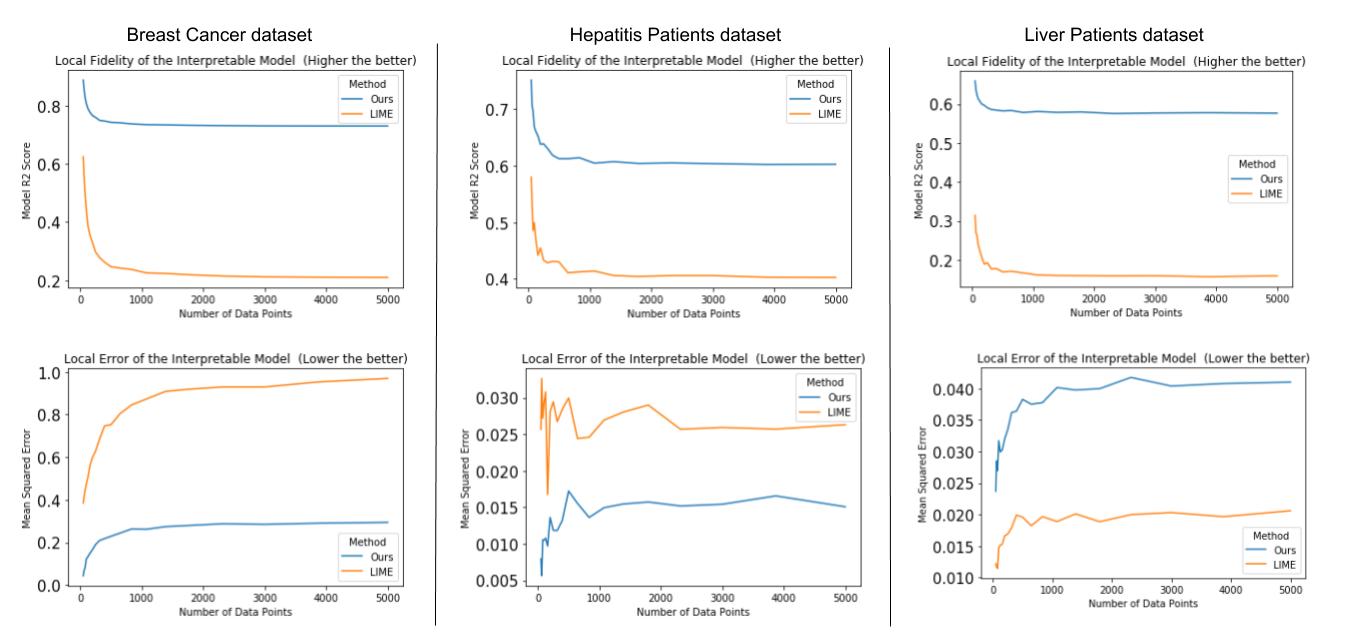}

\caption{Comparisons between our method and LIME \cite{lime} for local fidelity. The first row corresponds to the R2 score of the local surrogate model, the second row corresponds to the mean of MSE for all the points in the test set. In all cases, the number of data points for the local surrogate model is varied in log scale.} \label{fig:localfidelity}
\end{figure}
Currently, there exists no suitable metric for proper comparison of the two different interpretable models. Since our focus is on local fidelity and stability, we define and employ suitable metrics for the two issues of focus. For local fidelity, the local surrogate model should fit the global black box model locally. To test this, we compute the $R^2$ score of the local surrogate model using the results from the black box model as the ground truth. This tells us how good the model has fit on the generated data points. We compute the mean $R^2$ scores considering all the points in the test set. We also test the local model for fidelity by computing the mean squared error (MSE) between the local model prediction and the black box model prediction for the instance of interest that is to be explained. We again compute mean MSE considering all the points in the test set for the three respective datasets. Additionally, to study the effects of the dataset size used for the local surrogate model, we vary the number of generated data points used for training the local surrogate model. The results for the local fidelity experiments are shown in Figure \ref{fig:localfidelity}. It can be seen that in terms of both metrics, ALIME clearly outperforms LIME by providing a better local fit. The results seem consistent across the three datasets.  
\begin{figure}
\includegraphics[width=\textwidth]{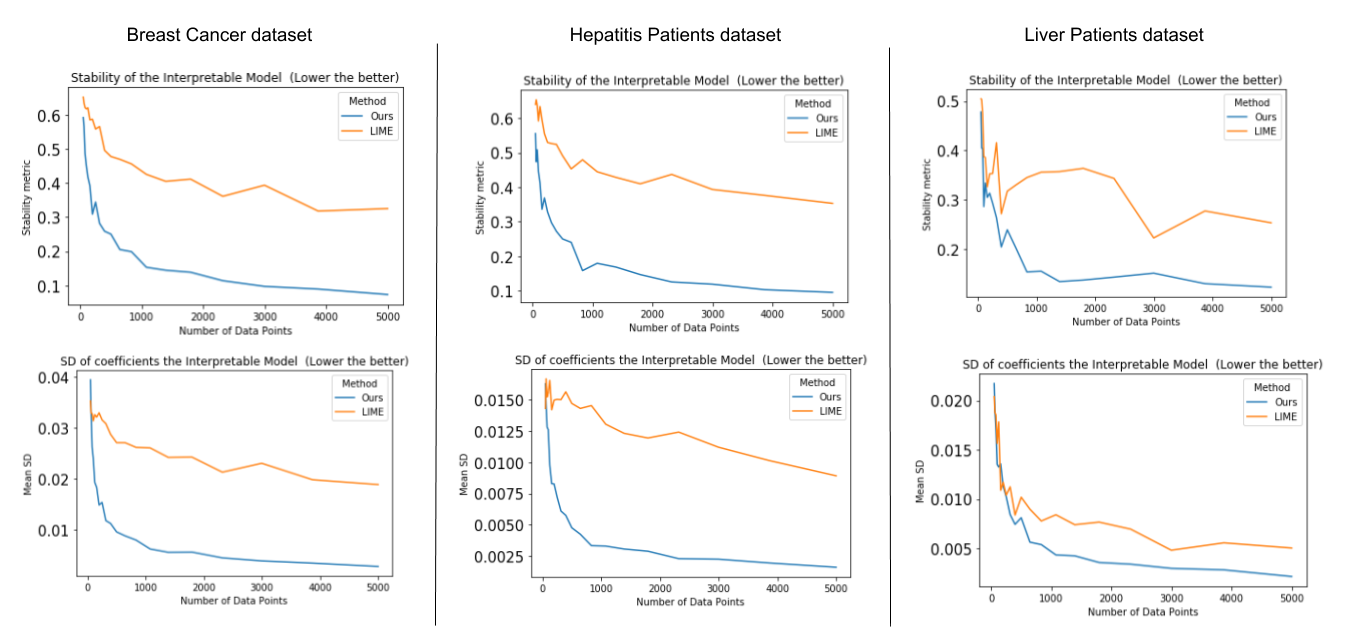}
\caption{Comparisons between our method and LIME \cite{lime} for stability. The first row corresponds to the mean stability metric of $10$ iterations for the same instance and for all the features, the second row corresponds to the mean standard deviations of $10$ iterations for the same instance and for all the features. The number of data points for the local surrogate model is varied in log scale.} \label{fig:stability}
\end{figure}

It is even more difficult to define a suitable metric for the interpretable model stability. Since the explanations are based on the surrogate model's coefficients, we can compare the change in the values of the coefficients for multiple iterations. Randomly selecting a particular instance from the test set, we run both LIME and ALIME for $10$ iterations. Because of their nature, both methods sample different set of points at every iteration. Because of the different dataset used at every iteration, the coefficients' values change. As a measure for stability, one of the things we compare is the standard deviations of the coefficients. For each feature, we first compute the standard deviation across $10$ iterations and then compute the average of standard deviations of all the features. We also compute the ratio of standard deviation to mean as another stability metric. The division by mean serves as normalization, since the coefficients tend to have varied ranges. Similar to the above, we study the effects of the size of the dataset used for the local surrogate model, and vary the number of generated data points used for the local interpretable model. For every size, we compute the average of the aforementioned two stability metrics across all features. The results are plotted in Figure \ref{fig:stability}. It should be noted that we only consider the absolute values of coefficients while computing the means and standard deviations. Again, it can be seen that the ALIME outperforms LIME in terms of both the metrics and across the three datasets.


\section{Conclusion}

In this paper, we proposed a novel approach for explaining the model predictions for tabular data. We built upon the LIME \cite{lime} framework and proposed modifications by employing an autoencoder as the weighting function to improve both stability and local fidelity. With the help of extensive experiments, we showed that our method yields in better stability as well as local fidelity. Although we have shown the results empirically, a more thorough analysis is needed to substantiate the improvements. In future, we would work on performing a theoretical analysis and also exhaustive empirical analysis spanning different types of data.

%
%
\bibliographystyle{splncs04}
\bibliography{alime}

\end{document}